\pdfoutput=1

\documentclass[11pt]{article}
\usepackage{authblk}
\usepackage[]{hyperref}
\usepackage{tikz}

\usepackage{acl}
\usepackage{float}
\restylefloat{table}

\usepackage{multirow}
\usepackage{enumitem}
\usepackage{times}
\usepackage{latexsym}
\usepackage{amsmath}
\usepackage{stmaryrd}
\usepackage{mathrsfs}
\usepackage{setspace}
\usepackage{times}
\usepackage{standalone}
\usepackage{bm}
\usepackage{mathtools} 
\usepackage{amssymb} 
\usepackage{array}
\usepackage{soul}
\usepackage{xcolor} 
\usepackage{tabularx}

\definecolor{customorange}{RGB}{224, 131, 24}
\definecolor{customred}{RGB}{172, 23, 12}
\definecolor{customgreen}{RGB}{74, 133, 50}

\DeclareRobustCommand{\inst}[1]{{\textcolor{customorange}{#1}}}
\DeclareRobustCommand{\loc}[1]{{\textcolor{customred}{#1}}}
\DeclareRobustCommand{\dat}[1]{{\textcolor{customgreen}{#1}}}

\usetikzlibrary{shapes.multipart}

\usetikzlibrary{fadings,calc,shapes.misc,arrows,decorations.pathmorphing,backgrounds,positioning,fit,petri}
\tikzset{font=\small}
\tikzstyle{flow} = [rectangle, rounded corners, minimum width=5cm, minimum height=1cm,text centered, draw=black]
\tikzstyle{arrow} = [thick,->,>=stealth]
\tikzstyle{phrase} = [rectangle, rounded corners, minimum height=0.4cm,text centered, draw=black, inner sep=2pt]
\tikzstyle{line} = [semithick,-,>=stealth]
\tikzstyle{word} = [rectangle, text centered, inner sep=0pt]
\tikzstyle{edition} = [rectangle, rounded corners, draw=black, dashed, inner sep=1pt, minimum width=\textwidth]
\tikzset{latin1/.style={node distance=0.6cm,}}
\tikzset{english1/.style={node distance=0.35cm,}}
\tikzset{german1/.style={node distance=0.25cm,}}
\tikzset{latin2/.style={node distance=0.6cm,}}
\tikzset{english2/.style={node distance=0.45cm,}}
\tikzset{german2/.style={node distance=0.25cm,}}

\newcolumntype{C}{>{\centering\arraybackslash}p{.8cm}}

\usepackage{etoolbox}

\usepackage[T2A,T1]{fontenc}

\usepackage[utf8]{inputenc}
\usepackage{booktabs}
\usepackage{newunicodechar}  

\usepackage{microtype}
\usepackage{graphicx}

\usepackage[russian,english]{babel}
\usepackage{substitutefont}

\substitutefont{T2A}{\familydefault}{Tempora-TLF}

\usepackage{amsmath}

\DeclareMathOperator{\inside}{in}
\DeclareMathOperator{\outside}{out}
\DeclareMathOperator{\final}{final}
\DeclareMathOperator{\latin}{Latin}

\DeclareMathOperator{\candidates}{candidates}

\newunicodechar{🐫}{\includegraphics[height=1.25\fontcharht\font`A]{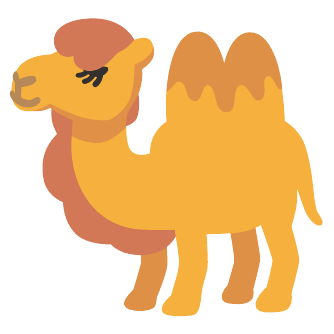}}

\title{CaMEL: Case Marker Extraction without Labels 🐫}

\author[*]{Leonie Weissweiler}
\author[$\dag$*]{Valentin Hofmann}
\author[*]{Masoud Jalili Sabet}
\author[*]{Hinrich Sch\"utze}

\affil[*]{Center for Information and Language Processing, LMU Munich}
\affil[$\dag$]{Faculty of Linguistics, University of Oxford \protect\\ \texttt{\{weissweiler,masoud\}@cis.lmu.de} \protect\\ \texttt{valentin.hofmann@ling-phil.ox.ac.uk}}

\def\figref#1{Figure~\ref{fig:#1}}
\def\figlabel#1{\label{fig:#1}\label{p:#1}}

\def\tabref#1{Table~\ref{tab:#1}}

\def\tablabel#1{\label{tab:#1}\label{p:#1}}

\def\secref#1{\S\ref{sec:#1}}
\def\seclabel#1{\label{sec:#1}}
\def\eqref#1{Eq.~\ref{eqn:#1}}

\long\def\eat#1{\ignorespaces}                  

\begin{document}
\maketitle
\begin{abstract}
We introduce \textbf{CaMEL} (\textbf{Ca}se \textbf{M}arker \textbf{E}xtraction without \textbf{L}abels), a novel and
challenging task in computational morphology that is especially relevant
for low-resource languages. We propose a first model for CaMEL that uses 
a massively multilingual corpus
to extract case markers in 83 languages
based only on a noun phrase chunker and an alignment system. To evaluate
CaMEL, we automatically construct a silver standard from UniMorph. 
The case markers extracted by our model can be used to detect and visualise
similarities and differences between the case systems of different
languages as well as to annotate fine-grained deep cases in languages in
which they are not overtly marked.
\end{abstract}

\section{Introduction}
\seclabel{intro}
What is a case?  Linguistic scholarship has shown that there
is an intimate relationship between morphological case
marking on the one hand and semantic content on the other
(see \citet{Blake.1994} and \citet{Grimm.2011} for
overviews).  For example, the Latin case
marker \textit{-ibus}\footnote{In this paper, we
use \textit{italic} when talking about case markers as
morphemes in a linguistic context and \texttt{monospace}
(accompanied by \texttt{\$} to mark word boundaries) when
talking about case markers in the context of our
model. Transliterations of Cyrillic examples are given after
slashes.} (Ablative or Dative Plural) can express the semantic
category of location. It has been observed that there is a
small number of such semantic categories frequently found
cross-linguistically \citep{Fillmore.1968, Jakobson.1984},
which are variously called \textit{case roles} or \textit{deep
cases}. Semiotically, the described situation is complicated
by the fact that the relationship between case markers and
expressed semantic categories is seldom isomorphic, i.e.,
there is both \textit{case polysemy} (one case, several
meanings) and
\textit{case homonymy}
or  \textit{case syncretism} (several cases, one marker) \citep{Baerman.2009}. As illustrated in Figure \ref{fig:example}, 
the Latin Ablative marker \textit{-ibus} can express
the semantic category of instrument besides location (case polysemy), and it is also the marker
of the Dative Plural expressing a recipient (case syncretism). In addition, there is 
\textit{case synonymy} (one case, several markers), which
further complicates morphosemiotics; e.g.,  in Latin, \textit{-is} is an alternative marker of the Ablative Plural.

\begin{figure}
  \centering
  \begin{tikzpicture}[level distance=6em]
  
    \renewcommand*{\arraystretch}{1.2}
    \node[rounded corners, draw,inner sep=0pt, line width=0.25mm] {
    \begin{tabular}{C|C|C}
    \multicolumn{3}{c}{\textit{-ibus}} \\
    \hline
    \multicolumn{2}{c|}{\textsc{Abl}} & \textsc{Dat} \\
    \hline
     \inst{I} & \loc{L} & \dat{R}\\
    \end{tabular}}
    child { node[rounded corners, draw,inner sep=0pt, line width=0.25mm] {
      \begin{tabular}{C}
        \foreignlanguage{russian}{\textit{-ами}}\\
        \hline
        \textsc{Inst} \\
        \hline
        \inst{I}
      \end{tabular}}}
       child { node[rounded corners, draw,inner sep=0pt, line width=0.25mm] {
        \begin{tabular}{C}
          \foreignlanguage{russian}{\textit{-ах}}\\
          \hline
          \textsc{Loc} \\
          \hline
          \loc{L}
        \end{tabular}}}
       child { node[rounded corners, draw,inner sep=0pt, line width=0.25mm] {
      \begin{tabular}{C}
        \foreignlanguage{russian}{\textit{-ам}}\\
        \hline
        \textsc{Dat}\\
        \hline
        \dat{R}
      \end{tabular}}};
    
    \end{tikzpicture}
  \caption[]{Morpho-semiotic foundation of this study. The Latin case marker \textit{-ibus} is used for
  both the Ablative (\textsc{Abl}) and the Dative (\textsc{Dat}), which in turn express the three semantic categories of instrument (\textcolor{customorange}{I}), 
  location (\textcolor{customred}{L}), and recipient (\textcolor{customgreen}{R}). This is an example of both case polysemy
(one case: \textsc{Abl}, several meanings: \textcolor{customorange}{I} and \textcolor{customred}{L}) and 
case syncretism (several cases: \textsc{Abl} and \textsc{Dat}, one marker: \textit{-ibus}).
Russian, on the other hand,
        has an isomorphic relationship between 
            Instrumental (\textsc{Inst}), Locative (\textsc{Loc)}, and Dative (\textsc{Dat}), 
            the case markers corresponding to them
  (\foreignlanguage{russian}{\textit{-ами}}/\textit{-ami},
  \foreignlanguage{russian}{\textit{-ах}}/\textit{-ax},
  \foreignlanguage{russian}{\textit{-ам}}/\textit{-am}), and the expressed semantic categories
            (\textcolor{customorange}{I}, \textcolor{customred}{L}, \textcolor{customgreen}{R}).}
  \figlabel{example} 
\end{figure}
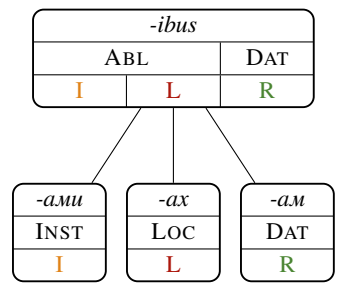

The key idea of this paper is to detect such complex correspondences between case markers 
and expressed semantic categories
in an automated way.
Specifically, we build on prior work by \citet{cysouw2014}, 
who lays the theoretical foundation for our study by showing 
that deep cases can be induced from cross-linguistic usage patterns 
of case markers. As opposed to Latin, Russian has separate cases (with separate case markers) for
the semantic categories of instrument (\foreignlanguage{russian}{\textit{-ами}}/\textit{-ami}), 
location (\foreignlanguage{russian}{\textit{-ах}}/\textit{-ax}),
and recipient (\foreignlanguage{russian}{\textit{-ам}}/\textit{-am}). Thus,
knowing the Russian case marker corresponding to Latin \textit{-ibus} 
reduces the uncertainty about the expressed case role
(Figure \ref{fig:example}).
This reduction of uncertainty  can be particularly
helpful in a low-resource setting where other means of analysis are unavailable. 

In this work, we rely on
the Parallel Bible Corpus
(PBC; \citealp{mayer-cysouw-2014-creating}), a massively multilingual corpus, 
to investigate
the relationship between surface cases and their deep
meanings cross-linguistically.
To put our idea into practice, we require an exhaustive set of case markers as well as a set of parallel noun phrases (NPs) that we can further 
analyze with respect to deep cases using the set of case markers. Both requirements pose a serious challenge 
for languages with limited available resources. We therefore introduce \textbf{CaMEL} (\textbf{Ca}se \textbf{M}arker \textbf{E}xtraction without \textbf{L}abels), a novel and
challenging task of finding case markers using only (i) a
highly parallel corpus covering many languages, (ii) a noun
phrase chunker for English, and (iii) word-level pre-computed alignments across languages.

Our work uses the parallel nature of the data in two ways.

First, we leverage the word-level alignments for the initial step of
our pipeline, i.e., the marking of NPs in all languages (even
where no noun phrase chunker is available). To do so, we
mark NPs in 23 different English versions of the Bible 
and project these annotations from each English to each non-English version using the word-level alignments,
resulting in parallel NPs that express the same semantic content across 83 languages.
Based on the projected annotations, 
we leverage the frequencies of potential case markers inside and outside of NPs
as a filter to distinguish case markers from lexical morphemes and other grammatical morphemes 
typically found outside of NPs. 

Second, we leverage the alignments for a fine-grained analysis of the semantic correspondences between case systems of different languages.

We make three main \textbf{contributions}.
\begin{itemize}[leftmargin=*]
\item We define \textbf{CaMEL}
(\textbf{Ca}se \textbf{M}arker \textbf{E}xtraction
without \textbf{L}abels), a new and challenging task with
high potential for automated linguistic analysis of cases
and their meanings in a multilingual setting.
\item We propose a simple method for CaMEL that is efficient, requires no training, and generalises well to low-resource languages.
\item We automatically construct a silver standard based on
human-annotated data and evaluate our method against it,
achieving an F1 of 45\%.
\end{itemize}
To foster future research on CaMEL, we make the silver standard, our code, and the extracted case
markers publicly available\footnote{\url{https://github.com/LeonieWeissweiler/CaMEL}}.
 
\section{Related Work}

Unsupervised morphology induction has long been
a topic of central interest in natural language processing
\citep{yarowsky-wicentowski-2000-minimally, goldsmith-2001-unsupervised, schone-jurafsky-2001-knowledge, creutz-lagus-2002-unsupervised, hammarstrom-borin-2011-unsupervised}. Recently, unsupervised inflectional
paradigm learning has attracted particular interest in the research community \citep{erdmann-etal-2020-paradigm, jin-etal-2020-unsupervised}, 
reflected also by a shared task devoted to the issue \citep{kann-etal-2020-sigmorphon}.
Our work markedly differs from this line of work in that we are operating
on the level of case markers, not full paradigms, and
in that we are inducing morphological structure 
in a massively multilingual setting. 

There also have been studies 
on extracting grammatical information from text
by using dependency parsers \citep{chaudhary-etal-2020-automatic, pratapa-etal-2021-evaluating} 
and
automatically glossing text \citep{zhao-etal-2020-automatic, samardzic-etal-2015-automatic} 
as well as compiling full morphological paradigms from it \citep{moeller-etal-2020-igt2p}.
By contrast, our method is independent of such annotation schemata, and it is also simpler as it does not aim 
at generating full grammatical or morphological descriptions
of the languages examined.
There has been cross-lingual work in computational morphology before \citep{snyder-barzilay-2008-unsupervised, cotterell-heigold-2017-cross, malaviya-etal-2018-neural}, but not
with the objective of inducing inflectional case markers.

Methodologically, our work is most closely related to 
the SuperPivot model presented by \citet{asgari-schutze-2017-past},
who investigate the typology of tense in 1,000 languages
from the Parallel Bible Corpus
(PBC; \citealp{mayer-cysouw-2014-creating})
by projecting tense information from languages 
that overtly mark it to languages that do not.
Based on this, \citet{asgari-schutze-2017-past} perform a typological analysis of tense systems in which they use different combinations of tense markers to further divide a single tense in any given language.
Our work differs in a number of important ways. 
First, we do not manually select a feature to investigate but model all features in our chosen sphere of interest (i.e., case) at once. Furthermore, we have access to word-level rather than verse-level alignments and can thus make statements at
 a more detailed resolution (i.e., about individual NPs).
Finally, we extract features not only for a small selection of pivot languages, but even for languages that do not mark case 
``non-overtly'', i.e., in a way that deviates to a large
degree from a
simple 1--1 mapping (see discussion in \secref{intro}).

\section{Linguistic Background}

There is ongoing discussion in linguistic typology about the
extent to which syntactic categories are shared and can be
compared between the world's languages
(see \citet{Hartmann.2014} for an overview).
While this issue is
far from being
settled, there is a general consensus that (while not being a
language universal) there is a core of semantic categories
that are systematically found cross-linguistically, and that
are expressed as morphosyntactic case in many
languages. Here, we adopt this assumption without any
theoretical commitment, drawing upon a minimal set of deep
cases detailed in Table \ref{table:cases}.  The set is
loosely based on the classical approach presented
by \citet{Fillmore.1968}.

Going beyond deep cases, \citet{cysouw2014} envisages a more fine-grained analysis of what is conventionally clustered in a deep case or semantic role.
Briefly summarised, the theoretical concept is this: if every language has a slightly different case system, with enough languages it should be possible to divide and cluster NPs at any desired level of granularity, 
from the conventional case system down to a specific usage of a particular verb in conjunction with only a small set of nouns.
For example, the semantic category of location could be further subdivided into  
specific types of spatial relationships such as `within', `over' and `under'.
Taken together, it would then be possible to perform theory-agnostic typological analysis 
of case-like systems across truly divergent and low-resource languages by simply describing any language's case system in terms of its clustering of very fine-grained semantic roles into larger systems that are overtly marked.

The approach sketched in the last paragraph 
is not limited to case systems but has been applied to person marking \citep{cysouw2008building}, the causative/inchoative alternation \citep{cysouw2010semantic}, and motion verbs \citep{walchli2012lexical}. The variety of linguistic 
application areas highlights the 
potential of developing methods that are much more automated
than the work of Cysouw and collaborators.
While we stay at the level of traditional deep cases in this paper, we hope to 
be able to extend our method into the direction of a more general analysis tool in the future.

The remainder of the paper is structured as follows. Section \ref{sec:method} describes our method in detail.
Section \ref{sec:eval} gives an overview of our results. Finally, Section \ref{sec:explor}
presents two exploratory analyses.

\begin{table*}[]
  \centering
  \begin{tabularx}{\textwidth}{lll}
    \toprule
    Deep Case                         & Description                                                                & Example                           \\ \midrule
    \multicolumn{1}{l}{Nominative}   & \multicolumn{1}{l}{The subject of the sentence}                           & \underline{He} is the Messiah! \\
    \multicolumn{1}{l}{Genitive}     & \multicolumn{1}{l}{An entity that possesses another entity}                  & Are you the \underline{Judean People's} Front?\\
    \multicolumn{1}{l}{Recipient}    & \multicolumn{1}{l}{A sentient destination}                                & I gave the gourd \underline{to Brian}.        \\
    \multicolumn{1}{l}{Accusative}   & \multicolumn{1}{l}{The direct object of the sentence}                                     & Consider \underline{the lilies}.              \\
    \multicolumn{1}{l}{Locative}     & \multicolumn{1}{l}{The spatial or temporal position of an entity} & They haggle \underline{in the market}.        \\
    \multicolumn{1}{l}{Instrumental} & \multicolumn{1}{l}{The means by which an activity is carried out}         & The graffiti was written \underline{by hand}. \\
    \bottomrule
    \end{tabularx}
    \caption{Descriptions and examples for the deep cases distinguished in this paper, which loosely follow 
the core deep cases proposed in the classical approach of \citet{Fillmore.1968}. }
    \label{table:cases}
\end{table*}

\section{Methodology} \label{sec:method}

\subsection{Data}

We work with the subset of the PBC \citep{mayer-cysouw-2014-creating} for which the SimAlign alignment algorithm  \citep{jalili-sabet-etal-2020-simalign} is available, resulting in 87 languages for our analysis.
From the corpus, we only extract those verses that are available in all languages, thus providing for a relatively fair comparison, 
and remove Malagasy, Georgian, Breton, and Korean, as they have much lower coverage than the other languages. This leaves us with 83 languages and 6,045 verses as our dataset.
We also select 23 English versions from the PBC that cover
the same 6,045 verses. For each of the 6,045 verses, we then
compute
$83 \times 23 = 1909$ verse alignments:
83 (for each language) multiplied with 23 (for each English
version). In the following, we will describe the components of our pipeline (\figref{pipeline}).

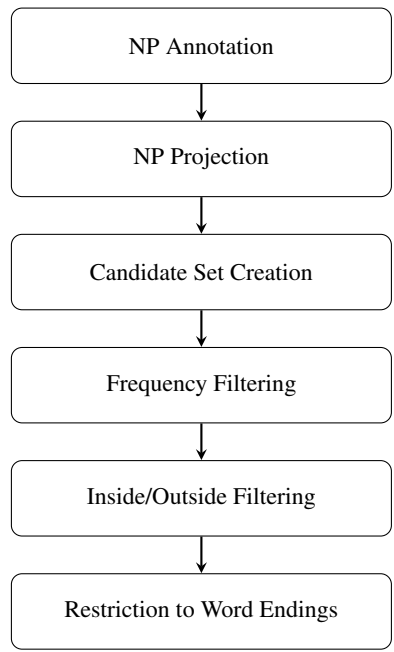
\begin{figure}
  \centering
 \begin{tikzpicture}[node distance=1.5cm]

   \node (annotation) [flow] {NP Annotation};
   \node (projection) [flow, below of=annotation] {NP Projection};
   \node (candidate) [flow, below of=projection] {Candidate Set Creation};
   \node (frequency) [flow, below of=candidate] {Frequency Filtering};
   \node (io) [flow, below of=frequency] {Inside/Outside Filtering};
   \node (ending) [flow, below of=io] {Restriction to Word Endings};
   \draw [arrow] (annotation) -- (projection);
   \draw [arrow] (projection) -- (candidate);
   \draw [arrow] (candidate) -- (frequency);
   \draw [arrow] (frequency) -- (io);
   \draw [arrow] (io) -- (ending);
   
   \end{tikzpicture}
  \caption{Overview of our pipeline.}
  \figlabel{pipeline}
\end{figure}                           

\subsection{NP Annotation}
Because our intermediate goal is to induce complete lists of case markers in all languages we cover, the first step is to restrict the scope of our search to NPs. 
We hope that this will allow us to retrieve case markers for
nouns and adjectives while disregarding verb endings that might otherwise have similar distributional properties.
As we are working with 83 languages, most of which are low-resource and lack high-quality 
 noun phrase chunkers, we first identify NPs in English using the spaCy noun phrase chunker \citep{spacy} and then project this annotation using the alignments to mark NPs in all other languages. The exception to this are German and Norwegian Bokmål, for which noun phrase 
 chunkers are available directly in spaCy.
Because both the spaCy noun phrase chunker and the alignments are prone to error, we make use of 23 distinct English versions of the Bible and mark the NPs in each of them with the goal of lessening the impact of noise.

\subsection{NP Projection}
\label{section:projection}
We project the NP annotation of a given English version to a second language using the alignments. Specifically, we find the NP in the target language by following the alignments from all words in the English NP while maintaining the word order of the 
target sentence. 
We treat each annotated version of the corpus resulting from the different English versions as a separate data source.
As an example, \figref{alignments} shows two English versions and
 the NP projections for Latin and German. While the
 alignments, particularly those from English to Latin, are
 not perfect, they result in complementary errors. The first
 wrongly aligns the first mention of \textit{pastor bonus},
 resulting in only \textit{pastor} being marked as an NP.
The second misses the alignment of \textit{life}
and \textit{animam}. In these two cases, the other alignment
corrects the error.

There are two major results from this process.

First, we obtain the set $N$ of
all NPs marked in English, each with all of its
translations in the other languages. An example of an entry in this set, taken
from  \figref{alignments}, would be \textit{the fine
shepherd, pastor bonus, der vortreffliche Hirte, ...},
while \textit{the fine shepherd, pastor, der vortreffliche
Hirte, ...} would be another, slightly defective, example.

Second, we obtain a pair of multisets, $W^l_{\inside}$ and
$W^l_{\outside}$, one for each language $l$.
$W^l_{\inside}$
(resp.\ $W^l_{\outside}$)
is the multiset
of all word tokens that appear inside (resp.\ outside) of NPs
of language $l$. In the following, we will use $M(w)$ to refer to the frequency of word $w$ in the multiset $M$.

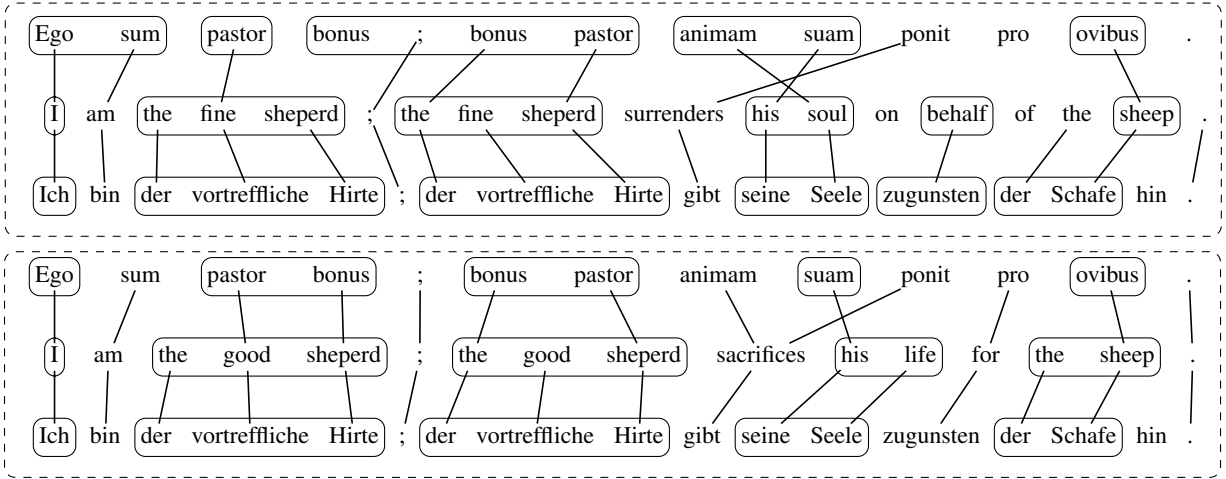
\begin{figure*}
  \begin{tikzpicture}[node distance=0.1cm and 0.2cm]

      \node (1ego) [latin1, word] {Ego\strut};
      \node (1sum) [latin1, word, right = of 1ego] {sum\strut};
    \node [latin1, phrase, fit=(1ego)(1sum)] {};
      \node (1pastor1) [latin1, word, right = of 1sum] {pastor\strut};
    \node [latin1, phrase, fit=(1pastor1)] {};
    \node (1bonus1) [latin1, word, right = of 1pastor1] {bonus\strut};
    \node (1semicolon1) [latin1, word, right = of 1bonus1] {;\strut};
      \node (1bonus2) [latin1, word, right = of 1semicolon1)] {bonus\strut};
      \node (1pastor2) [latin1, word, right = of 1bonus2] {pastor\strut};  
    \node (1bonuspastor)[latin1, phrase, fit=(1bonus1)(1pastor2)] {\strut};
      \node (1animam) [latin1, word, right = of 1pastor2] {animam\strut};
      \node (1suam) [latin1, word, right = of 1animam] {suam\strut};
    \node (1animamsuam) [latin1, phrase, fit=(1animam)(1suam)] {};
    \node (1ponit) [latin1, word, right = of 1suam] {ponit\strut};
    \node (1pro) [latin1, word, right = of 1ponit] {pro\strut};
      \node (1ovibus) [latin1, word, right = of 1pro] {ovibus\strut};
    \node (1ovibusp) [latin1, phrase, fit=(1ovibus)] {};
    \node (1dot1) [latin1, word, right = of 1ovibus] {.\strut};

    \node (1latin) [fit=(1ego)(1sum)(1pastor1)(1bonus1)(1semicolon1)(1bonuspastor)(1animamsuam)(1ponit)(1pro)(1ovibusp)(1dot1)] {};

      \node (1i) [english1, word, below= 0.7cm of 1ego] {I\strut};  
    \node (1ip)[english1, phrase, fit=(1i)] {};     
    \node (1am) [english1, word, right = of 1i] {am\strut};
      \node (1the1) [english1, word, right = of 1am] {the\strut};
      \node (1fine1) [english1, word, right = of 1the1] {fine\strut};
      \node (1sheperd1) [english1, word, right = of 1fine1] {sheperd\strut};
    \node (1thefinesheperd1)[english1, phrase, fit=(1the1)(1fine1)(1sheperd1)] {};
    \node (1semicolon2) [english1, word, right = of 1sheperd1] {;\strut};
      \node (1the2) [english1, word, right = of 1semicolon2] {the\strut};
      \node (1fine2) [english1, word, right = of 1the2] {fine\strut};
      \node (1sheperd2) [english1, word, right = of 1fine2] {sheperd\strut};
    \node (1thefinesheperd2)[english1, phrase, fit=(1the2)(1fine2)(1sheperd2)] {};
    \node (1surrenders) [english1, word, right = of 1sheperd2] {surrenders\strut};
      \node (1his) [english1, word, right = of 1surrenders] {his\strut};
      \node (1soul) [english1, word, right = of 1his] {soul\strut};  
    \node (1hissoul)[english1, phrase, fit=(1his)(1soul)] {};
    \node (1on) [english1, word, right = of 1soul] {on\strut};
      \node (1behalf) [english1, word, right = of 1on] {behalf\strut};
    \node (1behalfp)[english1, phrase, fit=(1behalf)] {};     
    \node (1of) [english1, word, right = of 1behalf] {of\strut};
    \node (1the3) [english1, word, right = of 1of] {the\strut};
      \node (1sheep) [english1, word, right = of 1the3] {sheep\strut};
    \node (1sheepp) [english1, phrase, fit=(1sheep)] {};
    \node (1dot2) [english1, word, right = of 1sheep] {.\strut};

    \node (1english) [below= of 1latin, fit=(1ip)(1am)(1thefinesheperd1)(1semicolon2)(1thefinesheperd2)(1surrenders)(1hissoul)(1on)(1behalfp)(1of)(1the3)(1sheepp)] {};

      \draw [line] (1ego)  -- (1i); 
      \draw [line] (1sum)  -- (1am); 
      \draw [line] (1pastor1)  -- (1fine1); 
      \draw [line] (1semicolon1)  -- (1semicolon2); 
      \draw [line] (1bonus2)  -- (1the2); 
      \draw [line] (1pastor2)  -- (1sheperd2); 
      \draw [line] (1animam)  -- (1soul); 
      \draw [line] (1suam)  -- (1his); 
      \draw [line] (1ponit)  -- (1surrenders); 
      \draw [line] (1ovibus)  -- (1sheep);

      \node (1ich) [german1, word, below = 0.7cm of 1i] {Ich\strut};  
    \node (1ichp)[german1, phrase, fit=(1ich)] {};  
    \node (1bin) [german1, word, right = of 1ich] {bin\strut};
      \node (1der1) [german1, word, right = of 1bin] {der\strut};
      \node (1vortreffliche1) [german1, word, right = of 1der1] {vortreffliche\strut};
      \node (1hirte1) [german1, word, right = of 1vortreffliche1] {Hirte\strut};
    \node (1dervortrefflichehirte1) [german1, phrase, fit=(1der1)(1vortreffliche1)(1hirte1)] {};
    \node (1semicolon3) [german1, word, right = of 1hirte1)] {;\strut};
      \node (1der2) [german1, word, right = of 1semicolon3] {der\strut};
      \node (1vortreffliche2) [german1, word, right = of 1der2] {vortreffliche\strut};
      \node (1hirte2) [german1, word, right = of 1vortreffliche2] {Hirte\strut};  
    \node (1dervortrefflichehirte2) [german1, phrase, fit=(1der2)(1vortreffliche2)(1hirte2)] {};
    \node (1gibt) [german1, word, right = of 1hirte2] {gibt\strut};
      \node (1seine) [german1, word, right = of 1gibt] {seine\strut};
      \node (1seele) [german1, word, right = of 1seine] {Seele\strut};
    \node (1seineseele) [german1, phrase, fit=(1seine)(1seele)] {};
      \node (1zugunsten) [german1, word, right = of 1seele] {zugunsten\strut};
    \node (1zugunstenp) [german1, phrase, fit=(1zugunsten)] {};
      \node (1der3) [german1, word, right = of 1zugunsten] {der\strut};
      \node (1schafe) [german1, word, right = of 1der3] {Schafe\strut};
    \node (1derschafe) [german1, phrase, fit=(1der3)(1schafe)] {};
    \node (1hin) [german1, word, right = of 1schafe] {hin\strut};
    \node (1dot3) [german1, word, right = of 1hin] {.\strut}; 

    \node (1german) [below= of 1english, fit=(1ichp)(1bin)(1dervortrefflichehirte1)(1semicolon3)(1dervortrefflichehirte2)(1gibt)(1seineseele)(1zugunstenp)(1derschafe)(1hin)(1dot3)] {};
    
    \draw [line] (1i)  -- (1ich); 
    \draw [line] (1am)  -- (1bin); 
    \draw [line] (1the1)  -- (1der1); 
    \draw [line] (1fine1)  -- (1vortreffliche1); 
    \draw [line] (1sheperd1)  -- (1hirte1); 
    \draw [line] (1semicolon2)  -- (1semicolon3); 
    \draw [line] (1the2)  -- (1der2); 
    \draw [line] (1fine2)  -- (1vortreffliche2); 
    \draw [line] (1sheperd2)  -- (1hirte2); 
    \draw [line] (1surrenders)  -- (1gibt); 
    \draw [line] (1his)  -- (1seine); 
    \draw [line] (1soul)  -- (1seele); 
    \draw [line] (1behalf)  -- (1zugunsten); 
    \draw [line] (1the3)  -- (1der3); 
    \draw [line] (1sheep)  -- (1schafe); 
    \draw [line] (1dot2)  -- (1dot3); 

    \node (1block) [edition, fit=(1latin)(1english)(1german)] {};

      \node (2ego) [latin2, word, below= 0.7cm of 1ich] {Ego\strut};
    \node (2egop) [latin2, phrase, fit=(2ego)] {};
    \node (2sum) [latin2, word, right = of 2ego] {sum\strut};
      \node (2pastor1) [latin2, word, right = of 2sum] {pastor\strut};
      \node (2bonus1) [latin2, word, right = of 2pastor1] {bonus\strut};  
    \node (2pastorbonus) [latin2, phrase, fit=(2pastor1)(2bonus1)] {};
    \node (2semicolon1) [latin2, word, right = of 2bonus1] {;\strut};
      \node (2bonus2) [latin2, word, right = of 2semicolon1)] {bonus\strut};
      \node (2pastor2) [latin2, word, right = of 2bonus2] {pastor\strut};
    \node (2bonuspastor) [latin2, phrase, fit=(2bonus2)(2pastor2)] {};
    \node (2animam) [latin2, word, right = of 2pastor2] {animam\strut};
      \node (2suam) [latin2, word, right = of 2animam] {suam\strut};
    \node (2suamp) [latin2, phrase, fit = (2suam)] {};   
    \node (2ponit) [latin2, word, right = of 2suam] {ponit\strut};
    \node (2pro) [latin2, word, right = of 2ponit] {pro\strut};
      \node (2ovibus) [latin2, word, right = of 2pro] {ovibus\strut};
    \node (2ovibusp)[latin2, phrase, fit=(2ovibus)] {};
    \node (2dot1) [latin2, word, right = of 2ovibus] {.\strut};

    \node (2latin) [fit=(2egop)(2sum)(2pastorbonus)(2semicolon1)(2bonuspastor)(2animam)(2suamp)(2ponit)(2pro)(2ovibusp)(2dot1)]{};

      \node (2i) [english2, word, below = 0.7cm of 2ego] {I\strut};  
    \node (2ip) [english2, phrase, fit=(2i)] {};
    \node (2am) [english2, word, right = of 2i] {am\strut};
      \node (2the1) [english2, word, right = of 2am] {the\strut};
      \node (2good1) [english2, word, right = of 2the1] {good\strut};
      \node (2sheperd1) [english2, word, right = of 2good1] {sheperd\strut};
    \node (2thegoodsheperd1) [english2, phrase, fit=(2the1)(2good1)(2sheperd1)] {};
    \node (2semicolon2) [english2, word, right = of 2sheperd1)] {;\strut};
      \node (2the2) [english2, word, right = of 2semicolon2] {the\strut};
      \node (2good2) [english2, word, right = of 2the2] {good\strut};
      \node (2sheperd2) [english2, word, right = of 2good2] {sheperd\strut};  
    \node (2thegoodsheperd2) [english2, phrase, fit=(2the2)(2good2)(2sheperd2)] {};
    \node (2sacrifices) [english2, word, right = of 2sheperd2] {sacrifices\strut};
      \node (2his) [english2, word, right = of 2sacrifices] {his\strut};
      \node (2life) [english2, word, right = of 2his] {life\strut};
    \node (2hislife) [english2, phrase, fit=(2his)(2life)] {};
    \node (2for) [english2, word, right = of 2life] {for\strut};
      \node (2the3) [english2, word, right = of 2for] {the\strut};
      \node (2sheep) [english2, word, right = of 2the3] {sheep\strut};
    \node (2thesheep) [english2, phrase, fit=(2the3)(2sheep)] {};
    \node (2dot2) [english2, word, right = of 2sheep] {.\strut};

    \node (2english) [fit=(2ip)(2am)(2thegoodsheperd1)(2semicolon2)(2thegoodsheperd2)(2sacrifices)(2hislife)(2for)(2thesheep)(2dot2)] {};

    \draw [line] (2ego)  -- (2i); 
    \draw [line] (2sum)  -- (2am); 
    \draw [line] (2pastor1)  -- (2good1); 
    \draw [line] (2bonus1)  -- (2sheperd1); 
    \draw [line] (2semicolon1)  -- (2semicolon2); 
    \draw [line] (2bonus2)  -- (2the2); 
    \draw [line] (2pastor2)  -- (2sheperd2); 
    \draw [line] (2animam)  -- (2sacrifices); 
    \draw [line] (2suam)  -- (2his); 
    \draw [line] (2ponit)  -- (2sacrifices); 
    \draw [line] (2pro)  -- (2for); 
    \draw [line] (2ovibus)  -- (2sheep); 
    \draw [line] (2dot1)  -- (2dot2);

      \node (2ich) [german2, word, below = 0.7cm of 2i] {Ich\strut};  
    \node (2ichp) [german2, phrase, fit=(2ich)] {};     
    \node (2bin) [german2, word, right = of 2ich] {bin\strut};
      \node (2der1) [german2, word, right = of 2bin] {der\strut};
      \node (2vortreffliche1) [german2, word, right = of 2der1] {vortreffliche\strut};
      \node (2hirte1) [german2, word, right = of 2vortreffliche1] {Hirte\strut};  
    \node (2dervortrefflichehirte1) [german2, phrase, fit=(2der1)(2vortreffliche1)(2hirte1)] {};
    \node (2semicolon3) [german2, word, right = of 2hirte1)] {;\strut};
      \node (2der2) [german2, word, right = of 2semicolon3] {der\strut};
      \node (2vortreffliche2) [german2, word, right = of 2der2] {vortreffliche\strut};
      \node (2hirte2) [german2, word, right = of 2vortreffliche2] {Hirte\strut};  
    \node (2dervortrefflichehirte2) [german2, phrase, fit=(2der2)(2vortreffliche2)(2hirte2)] {};
    \node (2gibt) [german2, word, right = of 2hirte2] {gibt\strut};
      \node (2seine) [german2, word, right = of 2gibt] {seine\strut};
      \node (2seele) [german2, word, right = of 2seine] {Seele\strut};  
    \node (2seineseele) [german2, phrase, fit=(2seine)(2seele)] {};
    \node (2zugunsten) [german2, word, right = of 2seele] {zugunsten\strut};
      \node (2der3) [german2, word, right = of 2zugunsten] {der\strut};
      \node (2schafe) [german2, word, right = of 2der3] {Schafe\strut};  
    \node (2derschafe) [german2, phrase, fit=(2der3)(2schafe)] {};
    \node (2hin) [german2, word, right = of 2schafe] {hin\strut};
    \node (2dot3) [german2, word, right = of 2hin] {.\strut};   

    \node (2german) [fit=(2ichp)(2bin)(2dervortrefflichehirte1)(2semicolon3)(2dervortrefflichehirte2)(2gibt)(2seineseele)(2zugunsten)(2derschafe)(2hin)(2dot3)] {};

    \draw [line] (2i)  -- (2ich); 
    \draw [line] (2am)  -- (2bin); 
    \draw [line] (2the1)  -- (2der1); 
    \draw [line] (2good1)  -- (2vortreffliche1); 
    \draw [line] (2sheperd1)  -- (2hirte1); 
    \draw [line] (2semicolon2)  -- (2semicolon3); 
    \draw [line] (2the2)  -- (2der2); 
    \draw [line] (2good2)  -- (2vortreffliche2); 
    \draw [line] (2sheperd2)  -- (2hirte2); 
    \draw [line] (2sacrifices)  -- (2gibt); 
    \draw [line] (2his)  -- (2seine); 
    \draw [line] (2life)  -- (2seele); 
    \draw [line] (2for)  -- (2zugunsten); 
    \draw [line] (2the3)  -- (2der3); 
    \draw [line] (2sheep)  -- (2schafe); 
    \draw [line] (2dot2)  -- (2dot3); 

    \node (2block) [edition, below = of 1block, fit=(2latin)(2english)(2german)] {};

    \end{tikzpicture}
  \caption{Example of alignments and NP projections
(English to Latin and English to German)
with two
  different English versions (top and bottom).}
  \figlabel{alignments}
\end{figure*}

For each language, we want to 
remove false positives from the word types contained within NPs (which are an artefact of 
wrong alignments) by
using the frequency of each word type inside and
outside of NPs.

In principle, this could be done
by means of a POS tagger and concentrating on nouns,
adjectives, articles, prepositions, and postpositions, but as we do not have
access to a reliable POS tagger for most languages covered
here, we use the relative frequency
information gained from our NP annotations. More specifically, 
we assign each
word type $w \in W^l_{\inside} \cup W^l_{\outside}$ to
$I_l$ (the set of words for language $l$ that are NP-relevant) if $|W^l_{\inside}(w)| >
|W^l_{\outside}(w)|$, and to $O_l$
(the set of words for language $l$ that are not NP-relevant) otherwise.
This
enhances the robustness of our method against occasional misannotations:
for Latin, \textit{ovibus} `sheep', from our previous example, occurred
once outside an NP but 45 times inside and is now an element
of $I_{\latin}$, while \textit{intellegent} `they understand' occurred once inside an
NP but 22 times outside and is therefore an element of
$O_{\latin}$.

\subsection{Candidate Set Creation}
From each language, we create a set of candidate case
markers
$\candidates(w)$ for a word $w$ 
by collecting all character $n$-grams of any length
from $w$ that are also members of $I_l$.  We explicitly mark the word
boundaries with \texttt{\$} so that $n$-grams in the middle
of words are distinct from those at the edges.  For example,
candidates extracted from \textit{ovibus} would
be \texttt{\$ovi}, \texttt{ibus\$}, but
also \texttt{\$ovibus\$} and \texttt{i}.
Our first candidate set is computed as
\mbox{$C_1^l = \bigcup \{\candidates(w) \mid w \in I_l\}$}.

\subsection{Frequency Filtering}
We define $I_l(c)$ as the number of words in $I_l$ that contain the candidate $c$, and $O_l(c)$ analogously for $O_l$.
As a first step, we filter out all $n$-grams with a
frequency in $I_l$ lower than a threshold
$\theta$.\footnote{We set $\theta = 97$ based on grid search.}
This results in \mbox{$C_2^l = \{c \mid c \in C_1^l, I_l(c) \geq \theta \}$}.

\subsection{Inside/Outside Frequency Filtering}
\label{sec:filter}

For this step, we make use of the observation that case is a
property of nouns. Hence, a case marker is expected to occur much more
frequently within NPs. This will serve to distinguish the
case markers from verb inflection markers, which should
otherwise have similar distributional properties.
To implement this basic idea,
for each candidate $c$ in language $l$, we first construct the contingency
table shown in Table \ref{table:2x2mono}.

\begingroup
\renewcommand{\arraystretch}{1.5}
\begin{table}[]
  \centering
  \small
  \caption{Contingency table for candidate case marker $c$
  in language $l$ for inside/outside filtering.
A morphological marker that occurs significantly more often
  inside NPs than outside of NPs is
  likely to be a nominal case marker.
}

  \label{table:2x2mono}
  \begin{tabular}{@{}c|c@{\hspace{20pt}}c@{}}
                    & $c$           & $\neg c$                               \\ \hline
    NP            & $I_{l}(c)$ & $\sum_{c' \neq c \in C_2^{l}} I_{l}(c')$             \\
    $\neg$ NP & $O_{l}(c)$ & $\sum_{c' \neq c \in C_2^{l}} O_{l}(c')$ \\ 
    \end{tabular}
\end{table}
\endgroup

We use the table to test whether a candidate is more or less
likely to appear inside NPs by comparing the frequencies of
the candidate inside and outside NPs to those of all other
candidates.  Shown in the cells are the frequencies used for
the test for each candidate.  The columns correspond to the
frequency of the candidate in question versus all other
candidates while the rows distinguish the frequencies inside
versus outside NPs.  We carry out a Fisher's Exact
Test \citep{fisher1922} on this table, which gives us a
$p$-value and an odds ratio $r$.  $r < 1$ if the candidate
is more likely to occur outside an NP, and $r > 1$ if it is
more likely to occur inside. The $p$-value gives us a
confidence score to support this ratio (lower is better). We
keep for $C_{\final}^l$ only those candidates for which $p
< \phi$ and $r> \chi$.\footnote{We set $\phi = 0.08$
and $\chi = 0.34$ based on grid search.}  For
example, \texttt{ibus\$} makes it past this filter with
$p($\texttt{ibus\$}$) = 2.869 \cdot 10^{-6}$ and
$r($\texttt{ibus\$}$) = 1.915$ -- it is significant and it
occurs inside NPs more often than outside NPs.
In contrast,
\texttt{t\$} is
discarded as it has $p($\texttt{t\$}$) = 3.18 \cdot
10^{-149}$ and $r($\texttt{t\$}$) = 0.249$ -- it is
significant, but it has
been found to occur much more likely outside than inside
NPs.

\subsection{Restriction to Word Endings}

Suffixoidal inflection is cross-linguistically more common than
prefixoidal and infixoidal inflection \citep{Bauer.2019}. This is also reflected in our dataset, where not a single language
has prefixoidal or infixoidal inflection. We hence restrict the set of considered $n$-grams
to ones at the end of words.

\section{Evaluation of Retrieved Case Markers} \label{sec:eval}

We evaluate our method for case marker extraction
without labels
using a silver standard.

\subsection{Silver Standard}

As we are, to the best of our knowledge, the first to introduce this task, we cannot rely on an existing set of gold case markers for each language we cover. 
As most of the languages included are low-resource, reliable grammatical resources do not always exist, which makes the handcrafting of a gold standard difficult.
Therefore, and also to ensure relative comparability, we evaluate against a silver standard automatically created from the UniMorph (\citealt{sylak2016composition}, \citealt{kirov-etal-2018-unimorph}, \citealt{mccarthy-etal-2020-unimorph})\footnote{https://unimorph.github.io} dataset.
The UniMorph data consists of a list of paradigms, which we first filter by their POS tag, keeping only nouns and adjectives and filtering out verbs and adverbs. 
An example of a paradigm is given in Table \ref{table:silver}. While the Nominative Singular (left column) is included in addition to the inflected forms (middle column), the straightforward approach of extracting the suffixes of the inflected forms is not optimal for every language, as the Nominative Singular form can differ from the root.
We therefore proceed as follows. 

First, we form a multiset of all inflected forms. 
In our example, this would result in 
\{\textit{Abflug, Abfluges, Abflug, Abflug, Abflüge, Abflüge, Abflügen, Abflüge}\}\footnote{Notice that we do
not handle diacritics in any special way. This limitation is
owed to the exploratory nature of our work. We see
diacritics as an exciting problem for the future.}. 
Next, we iterate over this multiset, removing one word each time 
if it occurs only once. This is meant to make the algorithm more 
robust against outlier words which do not share a common base with 
the rest of the paradigm. We then extract the longest common prefix 
for the remaining elements.  We build a frequency list of these prefixes,
which in our example has only one element, \textit{Abfl},
with a frequency of 3. We take the most frequent element
from the frequency list and compare it to the Nominative
Singular, \textit{Abflug}.  Of these two candidates, we take
the longer one.  We thereby prioritise precision over recall
as roots that are too short quickly result in many different
suffixes that are too long, due to the high overall number
of paradigms.  Finally, we iterate over the inflected forms
again, extracting the suffix if the chosen root is a prefix,
which in our example yields one new suffix: \texttt{es\$},
as \textit{Abflüge} and \textit{Abflügen} are not prefixed
by \textit{Abflug}.  We examine the results for each
language and exclude the languages where either basic
knowledge of the language or common sense makes it apparent
that sets are much too large or too small, resulting in a
diverse set of 19 languages to evaluate our methods against.
We note that this process automatically excludes adpositions
and clitics, which is in line with our focus on suffixoidal
inflection (Section \ref{sec:filter}). We make our silver
standard publicly available.

\begingroup
\begin{table}[]
\small
   \renewcommand{\arraystretch}{1.2}
    \centering
    \begin{tabular}{l|l@{\hspace{0.2cm}}ll|l@{\hspace{0.05cm}}l@{\hspace{0.05cm}}l}
Nominative & \multicolumn{3}{c}{inflected}&\multicolumn{3}{|c}{unused}\\
Singular &        \multicolumn{3}{c}{forms}  &\multicolumn{3}{|c}{information}\\
 & base &&suffix&\multicolumn{3}{c}{}\\\hline
\multirow{8}{*}{\textcolor{customorange}{Abflug}}  & \textcolor{customred}{Abfl}&\textcolor{customorange}{ug}&   & \textcolor{gray}{N} & \textcolor{gray}{NOM} & \textcolor{gray}{SG} \\
      & \textcolor{customred}{Abfl}&\textcolor{customorange}{ug}&\textcolor{customgreen}{es} & \textcolor{gray}{N} & \textcolor{gray}{GEN} & \textcolor{gray}{SG} \\
      & \textcolor{customred}{Abfl}&\textcolor{customorange}{ug} &  & \textcolor{gray}{N} & \textcolor{gray}{DAT} & \textcolor{gray}{SG} \\
      & \textcolor{customred}{Abfl}&\textcolor{customorange}{ug}&    & \textcolor{gray}{N} & \textcolor{gray}{ACC} & \textcolor{gray}{SG} \\
      & \textcolor{customred}{Abfl}&üge&  & \textcolor{gray}{N} & \textcolor{gray}{NOM} & \textcolor{gray}{PL} \\
      & \textcolor{customred}{Abfl}&üge&  & \textcolor{gray}{N} & \textcolor{gray}{GEN} & \textcolor{gray}{PL} \\
      & \textcolor{customred}{Abfl}&ügen& & \textcolor{gray}{N} & \textcolor{gray}{DAT} & \textcolor{gray}{PL}\\
      & \textcolor{customred}{Abfl}&üge&  & \textcolor{gray}{N} & \textcolor{gray}{ACC} & \textcolor{gray}{PL}
    \end{tabular}
    \caption{Example of silver standard creation. Marked in
    orange is the Nominative Singular form, in red the base
    (``base'') as determined by the algorithm, and in green
    the only suffix (``suffix'') that is extracted from this paradigm. Additional, unused information in the UniMorph data is marked in grey.}
    \label{table:silver}
  \end{table}
\endgroup

\subsection{Results}
\label{subsec:results}

Our results are provided in \tabref{fresults}. We
observe that precision is higher, at times even
substantially, than recall for most languages contained in
the silver standard. Looking at Table \ref{table:latin} as
an example, we can see that low precision is mostly due to
retrieved case markers being  longer
(\foreignlanguage{russian}{\texttt{ение}}\texttt{\$}/\textit{enie})
or shorter
(\foreignlanguage{russian}{\texttt{й}}\texttt{\$}/\textit{j})
than the correct ones.  It is one of the main challenges
in this task to select the correct length of a case marker
from a series of substring candidates. The shorter
substrings will automatically be more frequent and often
correct, but this is not easily solved by a frequency
threshold, which excludes other correct candidates that are
naturally less frequent.  Additionally, we observe that some
recall errors are due to an incorrect length of $n$-grams in
the silver standard
(\foreignlanguage{russian}{\texttt{ьям}}/\textit{'jam}),
highlighting that this issue also exists in its creation
process, and suggesting that our performance might even
improve when measured against handcrafted data.

\begin{table}[]
  \centering
  \begin{tabularx}{\columnwidth}{l@{\hspace{75pt}}r@{\hspace{20pt}}r@{\hspace{20pt}}r}
    \toprule
    Language& P & R & F1 \\ \midrule
    Albanian  & .74     & .47  & .58 \\
    Belarusian  & .43   & .41 & .42\\
    Bengali & .50    & .40  & .44 \\
    Czech  & .50     & .58  & .54 \\
    German  & .54    & .47  & .50 \\
    Greek  & .67     & .19  & .30 \\
    Icelandic  & .83  & .31  & .45 \\
    Indonesian  & .31     & .42  & .36 \\
    Irish  & .42     & .30 & .35 \\
    Latin  & .65     & .56  & .60 \\
    Lithuanian  & .18     & .38  & .24 \\    
    Nynorsk  & .79     & .48 & .59 \\
    Bokmål  & .67     & .45  & .54 \\
    Polish  & .52     & .33  & .40 \\
    Russian  & .54     & .54  & .54 \\
    Slovenian  & .41     & .28  & .33 \\
    Swedish  & .68     & .25  & .36 \\   
    Ukrainian  & .45     & .48  & .47 \\ 
    \midrule
    Average  & .54     & .41  & .45 \\
    \bottomrule
    \end{tabularx}
    \caption{Precision (P), Recall (R) and F1
on the task of case
marker extraction without labels
for languages
    contained in our silver standard. Nynorsk and Bokmål are
    two varieties of Norwegian.}
    \tablabel{fresults}
  \end{table}

\begin{table*}[]
  \centering
  \begin{tabularx}{\textwidth}{@{}p{4.9cm}@{\hspace{15pt}}p{4.9cm}@{\hspace{15pt}}p{4.9cm}@{}}
  \toprule
  Intersection & Algorithm Only & Silver Standard Only \\ \midrule
  \foreignlanguage{russian}{у, я, ом, ого, о, в, ой, и, ми, ам, ей, ю, ы, ов, ых, а, м, х, ами} & \foreignlanguage{russian}{ий, ные, ое, ение, ии, го, ый, ка, ые, к, ки, ия, ние, й, ния, ие} & \foreignlanguage{russian}{ыми, ах, ев, ьям, ому, ья, н, ьях, ями, ям, е, ях, ьев, ем, ым, ьями} \\
  \textit{u, ja, om, ogo, o, v, oj, i, mi, am, ej, ju, y, ov, yx, a, m, x, ami} & \textit{ij, nye, oe, enie, ii, go, yj, ka, ye, k, ki, ija, nie, j, nija, ie} & \textit{ymi, ax, ev, 'jam, omu, 'ja, n, 'jax, jami, jam, e, jax, 'ev, em, ym, 'jami} \\ \bottomrule  
  \end{tabularx}
  \caption{The output of our algorithm for Russian compared to the
  silver standard. We show suffixes that occur in the
  intersection of algorithm output and silver standard (``Intersection''),  those that occur only in the algorithm
  output
(``Algorithm Only'')
and those that occur only in the silver standard (``Silver Standard Only'').
To allow for a clear and concise presentation, the table does not observe the convention of using \$ for
  boundaries.}

  \label{table:latin}

\end{table*}

\subsection{Ablation Study}
We conduct an ablation study to assess the effects of the different pipeline components.

\subsubsection{Evaluating NP Projection}
In order to evaluate how well our method of projecting NP annotation using alignments to languages without an available NP chunker (see Section \ref{section:projection}) works, we evaluate it against the monolingual spaCy chunkers for Norwegian Bokmål and German, which are the only available languages besides English.
We do not directly compare annotated spans but instead their
influence on our method as we have intentionally designed our pipeline to be robust to some noise.
As $I_l$, the set of words considered to be NP-relevant, 
is the essential output of the annotation projection, we compare two versions, 
the set as a result of direct NP chunking and the set as a result of our annotation procedure. Taking the former as the ground truth
for evaluating the latter
(assuming that the directly chunked set has superior quality), 
we observe an F1 of 88.5\,\% for German and 67.8\,\% for
Norwegian Bokmål. While these numbers seem low at first, the
fact that our overall F1 on Norwegian Bokmål (.54,
see \tabref{fresults}) is better than on German (.50) indicates that the later elements of the pipeline are to a certain extent 
robust against misclassification of NPs. 

\subsubsection{Ablating Pipeline Components}

\begin{figure}
\centering
\includegraphics[width=0.9\linewidth]{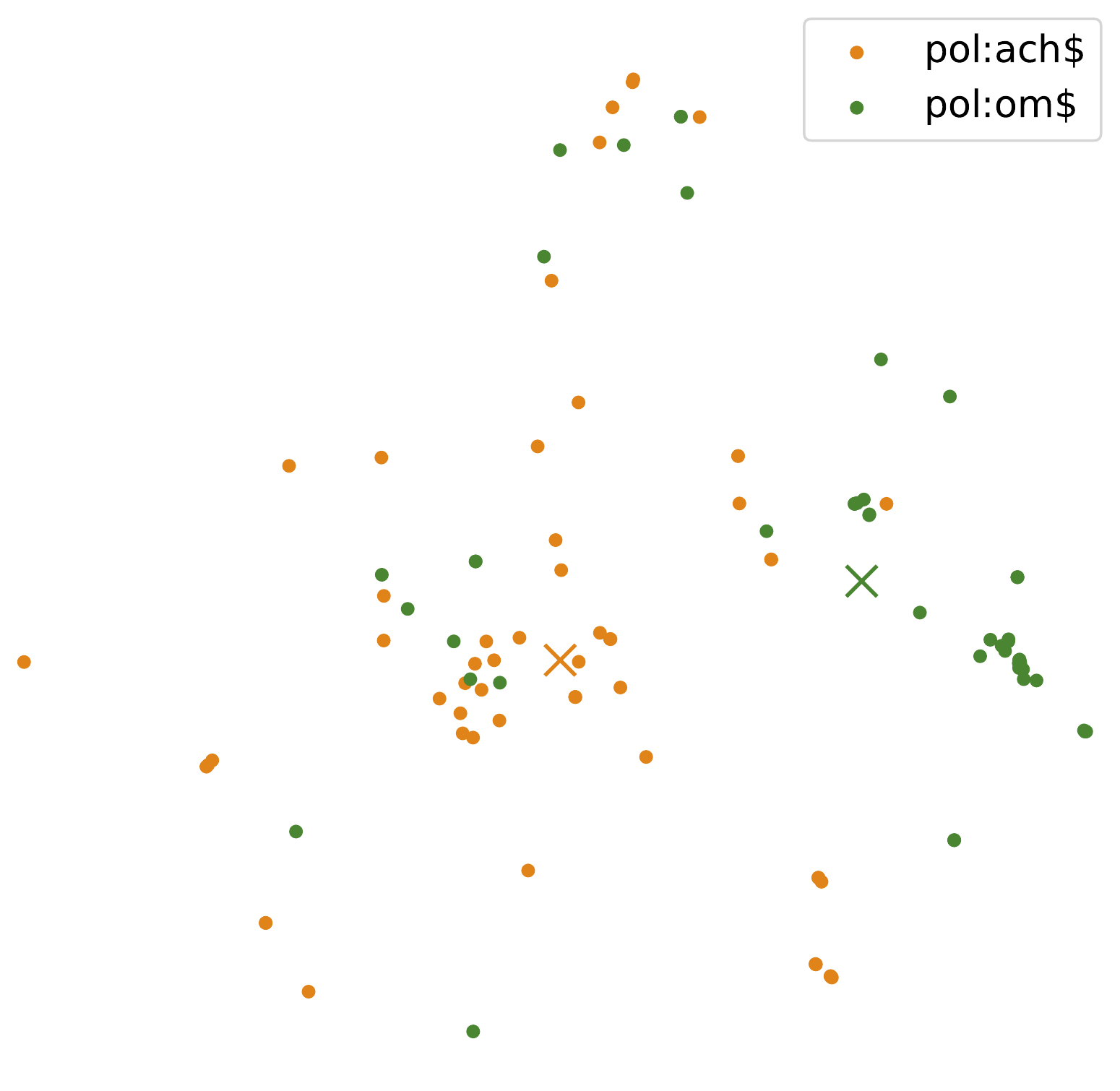}
  \caption{t-SNE plot of the contextual distribution of the
  Latin case marker \textit{-ibus} and the Polish case
  markers \textit{-ach} and \textit{-om}.  Outliers
  omitted.
The plot shows NPs who in Latin are marked with
the case marker
  \texttt{ibus\$}
and  in Polish either with 
\texttt{ach\$} (orange) or
   \texttt{om\$} (green). Centroids are marked with an X.  The
  plot shows that the Polish case markers exhibit a more
  fine-grained representation of the underlying semantic
  categories, which makes it possible to disambiguate the
  homonymous Latin case marker.}  \figlabel{ngrammap}
\end{figure}

We report the average Precison, Recall, and F1 across all languages in our silver standard without individual filtering components in Table \ref{table:ablation}.
Simple frequency filtering (see ``$\neg \theta$''), excluding
$n$-grams within words (see ``middle'') and at the beginning
of words (see ``beginning'') are all necessary for good
performance. Inside/outside filtering based on $p$-value is the most important
component of the pipeline (see ``$\neg \phi$''). Surprisingly, inside/outside
filtering based on odds ratio has almost no effect.

\begin{table}[]
  \centering
  \begin{tabular}{lllll}
    \toprule
    &ablation & P & R & F1 \\ \midrule\midrule
 our method  &&      &  &  \\
(\tabref{fresults})  &&    .54  &.41   &.45  \\\midrule
    &$\neg \theta$ & .11 & .59 & .16 \\
    &$\neg \phi$ & .00 & .00 & .00 \\
    &$\neg \chi$  & .53 & .41 & .44 \\
    &middle & .11 & .41  & .17 \\
    &beginning & .33 & .41 & .35  \\ \bottomrule
    \end{tabular}
\caption{Precision (P), Recall (R) and F1 averaged over all languages
on the task of case
marker extraction without labels
when each step of our pipeline is ablated.  $\neg \theta$: no Frequency Filtering; 
$\neg \phi$: no Inside/Outside Filterung based on $p$-value; $\neg \chi$: no Inside/Outside Filtering
based on odds ratio; middle: include middle of the word; beginning: include beginning of the word.}
\label{table:ablation}
\end{table}

\section{Exploratory Analyses} \label{sec:explor}

We can use our automatically extracted case markers,
in combination with the
parallel NPs that are extracted as part of the pipeline,
for   innovative linguistic analyses.
We present two examples in this section.

\subsection{Marking of Deep Cases}
First, we demonstrate how, given a parallel NP, the case
markers can be used to determine its deep case.  We return
to $N$ (see Section \ref{section:projection}), our set of
parallel NPs extracted from the PBC, and for a selected
subset of languages, group them by their combination of case
markers.
The basic idea is to infer an NP's (potentially very
fine-grained) deep case
by representing it as its
combination of case markers across languages.

For example, we can disambiguate the Latin case
marker \textit{-ibus} by looking at the different groups the
NPs containing it form with Russian case markers.  Recall
that \textit{-ibus} can express location, instrument, and
recipient and that   Russian expresses  these categories by
separate case markers:
\foreignlanguage{russian}{\textit{-ах}}/\textit{-ax} for
location, \foreignlanguage{russian}{\textit{-ами}}/\textit{-ami}
for instrument,
and \foreignlanguage{russian}{\textit{-ам}}/\textit{-am} for
recipient (see \figref{example}) --  all three of which have been retrieved by our
method.
Given a Latin NP marked by the ending \textit{-ibus}, the parallel NP in Russian
can help us determine its deep case. Thus, for
domibus, \foreignlanguage{russian}{дворцах}/\textit{dvorcax}
shows that the semantic category is location, i.e., `in the
houses'. For operibus
bonis, \foreignlanguage{russian}{добрыми
делами}/\textit{dobrymi delami} shows that the semantic
category is instrument, i.e., `through the good
deeds'. Finally, for patribus,
\foreignlanguage{russian}{предкам}/\textit{predkam} shows that the semantic category is a recipient, 
i.e., `for/to the parents'.

\subsection{Similarities between Case Markers}
We also demonstrate how we can use their distributional similarities over NPs to show how case markers that are similar in this respect correspond to similar combinations of deep cases.
We first generate an NP-word cooccurrence matrix over the NP
vocabulary of all languages in which each row, corresponding to
an inflected word firm $w$  in language $l$,
indicates which NPs (corresponding to columns) cooccur with 
$w$. in the parallel data.
We then reduce the dimensionality of the matrix by means of t-SNE \citep{van2008visualizing}, allowing us
to inspect systematic patterns with respect to the
``contexts'' in which certain case markers occur (where
``context'' refers to words the case marker is aligned to in
other languages, not words the case marker coccurs with in
its own language).
In a semiotic situation like the one shown in Figure \ref{fig:example}, this setup 
allows us to examine how the semantic region expressed by a certain homonymous case marker
in one language is split into more fine-grained regions in another language that 
distinguishes the semantic categories
that are lumped together by the case marker
(and which, if they are
at the right level of abstraction, can  correspond to  deep cases).

\figref{ngrammap} shows this scenario for the Latin Ablative
marker \textit{-ibus}.
It  corresponds to two distinct case markers in Polish, \textit{-ach} (\textsc{Loc}) and \textit{-om}
(\textsc{Dat}). The figure shows that the region occupied by Latin \textit{-ibus} splits into two 
distinct clusters in Polish, allowing us to visually determine which underlying case is expressed by the homonymous suffix \textit{-ibus}. This underscores the exploratory potential of our approach.

\section{Conclusion and Future Work}
We have introduced the new and challenging task of Case
Marker Extraction without Labels (CaMEL) and presented
a simple and efficient method that leverages
cross-lingual alignments and achieves an F1 of 45\% on 19
languages. We introduce an automatically created
silver standard to conduct our evaluation.  We have further demonstrated two ways in
which our retrieved case markers can be used for linguistic
analysis.

We see two potential avenues for future work.  The first is
the further improvement of case marker extraction.  The main
problem to tackle here is that of small sets of overlapping
substrings of which only one is the correct marker, and
developing some further measures by which they can be
distinguished.  Furthermore, it would be useful to find data
from more low-resource languages and languages that have
typological properties different from
the extensively studied large language families
(Indo-European, Turkic, Sino-Tibetan etc.). We could
then verify that our method performs well across languages
and attempt to expand our
silver standard to more languages while still ensuring
quality.  The second area is that of further automating the
analysis of deep case and case syncretism. Ideally, we would
develop a method that can distinguish  the different
possible reasons for
divergent case marking in languages, with the eventual
goal of creating a comprehensive overview of case and
declension systems for a large number of languages.

\section*{Acknowledgements}
This work was funded by the European Research
Council (\#740516). The second author was also supported by the German
Academic Scholarship Foundation and the Arts
and Humanities Research Council.
The third author was also supported by the German Federal Ministry of Education and Research (BMBF, Grant No. 01IS18036A).
We thank the reviewers for their extremely helpful comments.

\bibliography{anthology,custom,lit_valentin}

\end{document}